# Revisiting Causality Inference in Memory-less Transition Networks

Abbas Shojaee MD, CHDA[1,2] *
Isuru Ranasinghe MBChB, MMed, PhD [3]
Alireza Ani MD [4]

Author Affiliations
1. Pulmonary and Critical Care Section, Department of Internal Medicine, Yale University School of Medicine, New Haven, Connecticut
2. Center for Outcomes Research and Evaluation, Yale University, New Haven, Connecticut
3. Discipline of Medicine, The University of Adelaide, Adelaide, South Australia
4. Pegahsoft Co, Isfahan, Iran

* Author for Correspondence

Dr. Abbas Shojaee,
Pulmonary and Critical Care Section, Department of Internal Medicine
Yale University School of Medicine, New Haven, Connecticut 06510
Email: abbas.shojaee@yale.edu;
Phone: 203 747-6914
Fax:    203 785-3826

Disclosures
The authors declare that they have no relevant or material financial interests that relate to the research described in this paper.

# Abstract

Several methods exist to infer causal networks from massive volumes of observational data. However, almost all existing methods require a considerable length of time series data to capture cause and effect relationships. In contrast, memory-less transition networks or Markov Chain data, which refers to one-step transitions to and from an event, have not been explored for causality inference even though such data is widely available. We find that causal network can be inferred from characteristics of four unique distribution zones around each event. We call this 'composition of transitions' and show that cause, effect, and random events exhibit different behavior in their compositions. We applied machine learning models to learn these different behaviors and to infer causality. We name this new method "causality inference using composition of transitions" (CICT). To evaluate CICT, we used an administrative inpatient healthcare dataset to set up a network of patients' transitions between different diagnoses. We show that CICT is highly accurate in inferring whether the transition between a pair of events is causal or random and performs well in identifying the direction of causality in a bi-directional association.

# Introduction

Finding causal relationships is an important premise of scientific research including biomedical research. However, for identifying causal relationships in medicine – such as in the development of disease complications or treatment effects, we have continued to rely on the inference of experts and statistical correlations for such judgments. Causal relationships can be difficult to find and verify, in part because sufficient reliable data from clinical trials are sparse. Although massive amounts of observational data (such as billing data) exist that might yield relevant causal connections, little success has been achieved in Medicine in interrogating such data for identifying potential causal relationships.

Nonetheless, quantitative methods for detecting causal relations in observational data have been studied in different disciplines including physics, social networks, biology, genomics, epidemiology[1], economics and other disciplines. Granger causality, an important advancement in causality research, focuses on a linear relation between cause and effect and can be applied when information about a causative factor is not inseparably shared with the effect[2]. For nonlinear systems, different methods have been applied including nonlinear variations of Granger causality [3,4], techniques of state space reconstruction[2,5,6], conditional mutual information[7,8], recurrence plots[9,10] and information entropy transfer[11]. However, these methods require sufficiently large samples of long time series data to achieve reasonable results[12]. An important limiting factor in building these models is that causal inference methods make presumptions either about data structure (e.g. availability of a time series with sufficient length or consistent sequence of cause and effect) or about causal structure (e.g. being acyclic or non-recursive as in Bayesian networks). Such assumptions can be problematic for real-world data especially in complex and interconnected domains like medicine, biology, ecology, and finance. New methods to detect causal relationships that make minimal assumptions about data structure and causal structure are required to identify useful clinical insights using real-world observational data in an expedient, non-resource intensive manner.

In our study, we used an observational administrative healthcare dataset to evaluate whether causal relations can be inferred from the frequency of patients' transitions from one clinical condition to another (figure 1 A,B). Frequency data on observed events (phenomena) and transitions between them is inexpensive and commonly available in different scientific disciplines including health care. Such data can be used to setup a transition network by assuming each event as a node and aggregating all observed transitions between one node to another node as a connecting edge (figure 1 B,C). For example, for all patients who had pneumonia following an episode of influenza, one edge from the influenza node to the pneumonia node keeps the frequency of transition. Transition networks, known as stationary Markov chain, are ubiquitous in real-world data such as traffic data, the sequence of web clicks, message spreading, econometrics, ecology, weather prediction, and physics. Given the challenge of inference and prediction from Markov chain data, scientists have invented methods to meet this challenge, including community detection methods such as, Random walk and its offspring such as PageRank [13], Walktrap[14], and MapEquation[15]. However, these methods can be computationally expensive especially in large dense graphs and are not designed to infer causality.

## Causal Inference using Composition of Transitions (CICT)

We propose a novel method called **Causal Inference using Composition of Transitions (CICT)** for causal structure discovery in one-step transition Markov chain data. We demonstrate this method using observational data from the California State Healthcare Cost and Utilization Project State Inpatient Databases[16] which is an administrative healthcare dataset that records hospital admissions for various conditions. This dataset contains 15,047,413 hospital admissions among 3,966,603 patients who had two or more hospitalizations during 2005 to 2011. The primary admission diagnosis at each admission is coded using the International Classification of Disease 9th Edition Clinical Modification (ICD-9-CM) and represents the main clinical condition treated during the hospital encounter. We use this dataset to evaluate whether causal relations can be inferred from the frequency of patients' transitions from one clinical condition to another (Figure 1 A,B). We assumed that on a transition network the set of events before and after a cause have different stochastic compositions for an effect or random event. For example, events before and after myocardial infarction are different than before and after a respiratory infection. Also, we assumed that a transition from a cause to effect is an irreversible process in real-life. Such irreversibility should create observable asymmetry in transition rates for a cause-to-effect versus effect-to-cause. For example, the rate of transition from myocardial infarction (cause) to chronic heart failure (effect) should be higher than the reverse. To measure the differences in distribution of transitions, we considered two probabilities on each edge, given two nodes $i$: source and $j$: target, we defined confidence (Conf) and contribution (Contrib) as follow:

$$Conf_{ij} = \vec{ij}/i = probability\ of\ future\ transition\ to\ j\ conditioned\ on\ being\ in\ i$$

$$Contrib_{ij} = \vec{ij}/j = probability\ of\ a\ previous\ state\ of\ i\ conditioned\ on\ being\ in\ j$$

Thus, for each pair of nodes, we created two parameters for transition edge $\vec{ij}$ and two parameters for transition edge $\vec{ji}$. This results in 4 parameters for each pair of nodes. We called these four parameters first level features. Then we defined 2$^{nd}$ level parameters from 1$^{st}$ level features by normalizing $Conf_{ij}$ over $j$ ($Conf_{ij} * j/\sum j$) and $Contrib_{ij}$ over $i$ ($Contrib_{ij} * i/\sum i$). Next we calculated third level features by engineering 1$^{st}$ and 2$^{nd}$ level features borrowing ideas from graph networks and connected systems. For example, assuming edges are connections in an electrical circuit we calculated resistance ($abs(i-j)/\vec{ij}$), and taking the nodes and edges as a closed connected hydrolic system, we calculated output pressure ($(i-j) * \vec{ij}/i$). We created a total of 300 derivatives which top 8 important predictors of them are explained in the result section.

Next, for each node, $i$ we identified four unique zones of transition distributions as shown in figure 1-D. The first zone is the distribution of all confidences of node $i$ to other nodes. Zone 2, is the contribution of node $i$ to other nodes. Zone 3, is confidences of other nodes into node $i$. And Zone 4, is the contribution of other nodes into node $i$. To create a distribution for each parameter at each zone, we created a histogram by first sorting values of each parameter (for example Confidence in zone 1,4) on X axis and counting their frequencies in bins on Y axis. We called the distributions of transition in these 4 zones as the "composition of transitions". We hypothesized that cause, effect and random phenomena should exhibit different compositions. We named the specific pattern of composition for each type as its 'behavior'. Figure 1.E density graph shows the difference of behavior of a cause versus an effect and a random event in these zones. Red color shows distributions for "Rheumatoid Arthritis" which is a well-known cause of different clinical conditions and yellow represents "Syncope and Collapse" as a known effect of other conditions. We used "Pneumonia" as a random condition as it has the potential to affect patients with a wide range of previous medical conditions. The logarithmic scaled chart in figure 1.E shows that the median of distributions of transitions to and from Rheumatoid Arthritis (cause event) has a difference of two orders of magnitude with Pneumonia (random event). Syncope(effect) shows similarities to both.

Hence, to systematically measure the different behaviors of the cause, effect, and random events, we extracted statistics and moments of the distribution of confidences and contributions and their derivatives, in all the corresponding zones for each node. For example, we extracted mean, standard deviation, skewness, kurtosis, median absolute deviation[17-19] and L-moments[20,21] for confidences and contributions. Median absolute deviation and L-moments are measures of distribution that are more stable and less sensitive to outliers compared to standard measures, such as mean or standard deviation. We suggested that information about the behavior of source and target of a specific transition can provide clues to infer whether that transition is a cause, effect or random. Figure 1.F shows the 8 influencing distribution zones that we considered for each specific edge. We added the measures of distributions of these 8 zones as additional features to each specific transition edge. A total of 320 features were created to capture all possible facets of composition that would discriminate causes and causal transitions from others.

Hereafter we name it a **causal factor**, or briefly **causal** when a phenomenon is a precipitator, precursor or cause of a second phenomenon. We used unsupervised clustering methods and principal component analysis to evaluate whether the new features that we created can reveal an inherent grouping between known causal and non-causal conditions and transitions. Also, we trained classification machine learning models to learn differences between causal phenomenon versus effects and random occurrences. We used these models to predict causal relations, describe their predictive power and determine top predictors of causality. We chose an empirical approach to validate our results against well-known medical facts of proven causal relations (ground truth). The health domain provides a good testbed due to the availability of large-scale datasets and the benefit of well-established domain knowledge. Simultaneously, we avoided incorporating domain or design specific knowledge into the method (e.g. building models) to keep the findings as simple and generalizable as possible and to ensure applicability in other scientific domains. Also, we defined a minimum length model to show that even a minimal set of inputs, like simple one-step transitions frequencies, carry valuable information for causal inference.

## The Ground Truth

We require the ground truth to evaluate the results of unsupervised methods and to train supervised methods. Here the ground truth is existing knowledge about a sufficient set of relations between pairs of clinical conditions. To prepare this set we used Semantic MEDLINE Database (SemMedDB) from Semantic Knowledge Representation (SKR) project [22] that contains 82.2 million predicates between biomedical concepts extracted from all MEDLINE citations. We extracted a set of 267 causal relations from SemMedDB that we found a match for in our transition data. Two clinicians as subject matter experts verified the correctness of identified causal relations (Supplementary Table 1). Then we assigned the type of each relation to the corresponding transition on the graph. In addition to causal relations, a set of random relations is required for predictive models to learn the difference between causal and random relations. Accordingly, we chose a random sample of transitions from our transition graph, then, two subject matter experts manually tagged 267 relations as 'irrelevant-may coincide' denoting that a transition from the first clinical state to the second is most probably due to a random process and not a causal relation (Supplementary Table 2).

## Results of Causal Inference using Composition of Transitions (CICT)

Most modeling and machine learning studies, including in causality inference, are focused on building new models and proving the validity of modeling presumptions. Here, we use standard, well-established models and show that the features we created contain new information about causality that a standard model can learn. Accordingly, we are reporting the results of four experiments conducted using classification and clustering methods. We then evaluate the performance and results of the models to show that CICT can capture new facets of causality. Experiment I show the CICT power in discriminating causal transitions from random transitions. Experiment II shows how much CICT can predict the causal direction in a bi-directional association. The third experiment proves that CICT works independently of the event (state or phenomenon) identification process as long as it is consistent and reflective of real phenomena. The last experiment ensures the elimination of possible subjective errors in the validation phase by applying the trained model on the set of previously unknown and randomly selected transitions and evaluating the results. In all experiments after optimizing and validating the predictive model, we estimated the discrimination power of the models using the area under the receivers operating characteristics curve (AUC of ROC) as a surrogate of the amount of causal knowledge that CICT learns. In line with this, we used the predicted probability of classification as the measure of causality referred as 'measure' in tables. Also, we describe the most important predictors of causality and provide interpretation for causal behavior.

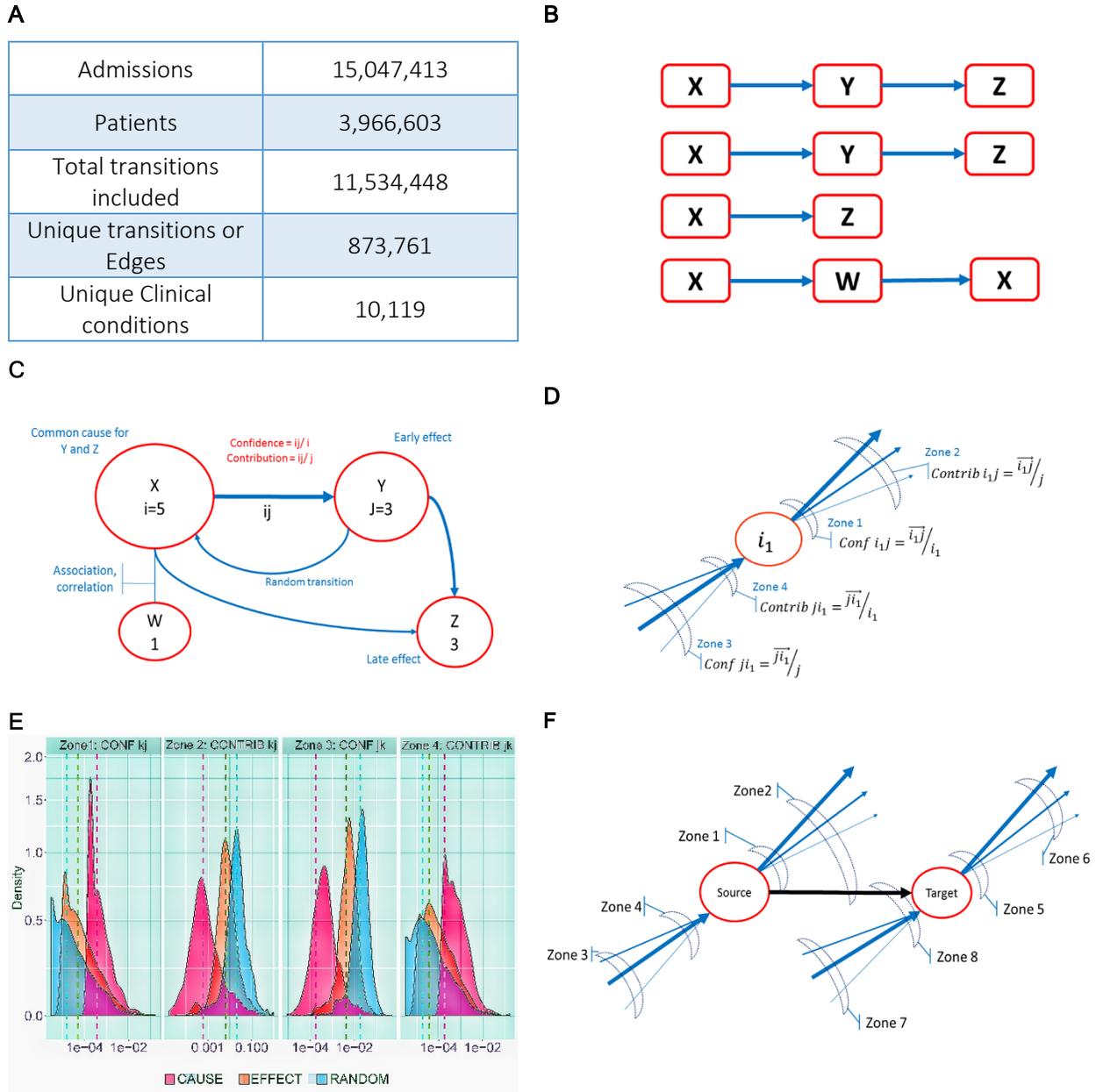

Figure 1: (A) Descriptive statistics on the data. (B) A set of transitions for four hypothetical patients. For example, the first patient is hospitalized with a principal diagnosis of condition X, and after a period is rehospitalized with condition Y and so forth. If we start merging similar transitions the result would be the transition graph shown in (C) Different types of transitions on a network. X is a common cause of Y and Z where Y is an early effect and Z is a late effect. X and W showed an association without an observable causal relation. Numbers represent hypothetical frequencies. (D) 4 zones that carry different distributional information. It is important to note that the 4 areas are not overlapping and contain different information. Here $i$ represents source and $j$ represents destination. (E) The log-scale density graph shows the different distributions in four distribution zones for a cause: Rheumatoid Arthritis(red), an effect: Syncope (yellow) and a random event: Pneumonia (blue). Doted lines show medians of correspondent density. (F) The 8 distribution zones that are identified above carry information relevant to the nature of the transition between source and destination. Zones 1,3,5,7 capture distribution of the parameters that are derived from Confidence calculation. Zones 2,4,6,8 capture distribution of the parameters that are derived from Contribution calculation.

# Results

## Experiment I: High accuracy in discriminating random transitions from causal associations

We used the ground truth to create a set of 267 random transitions and a set of 267 of causal relations. Then we split this 534 transition-set into a 75% training subset and 25% validation set. Next, we trained a random forest(RF) model with 10-fold cross-validation on the training subset to separate random relations from non-random relations. RF is a well-studied machine learning method that works well in nonlinear and complex problem domains by aggregating the collective result of multiple decision trees as its output. To ensure the stability of results we repeated training and testing 50 times. RF predictive model shows an average discrimination power of AUC= 0.916 with Mean Square Error = 0.074 and R2 = 0.699 on out of the bag samples. The model is well calibrated as evaluated by Hosmer-Lemeshow chi-square on 10 deciles of risk (Chi-square = 6.846, P-value = 0.553). (Figure 2 A,B,C). The fact that RF converges in just 3 decision trees of depth 5 means that separation of causal and random transitions can be inferred with a limited number of decisions over composition parameters. An area under curve greater than 0.9 for a model is considered as an excellent discrimination power. Top 10 relations predicted by the model are shown in Table 1.A. All the top 10 relations are well-known casual associations.

It is common knowledge that the right features are more important than technique sophistication in the performance of machine learning models. Moreover, right features can help unsupervised methods to group similar data points into clusters that reflect real classes. Figure 2 D shows two clusters, as identified by Partitioning Around Medoids (PAM)[23]. The denser cluster (cyan polygon) mostly embodies cause-effect transitions (blue triangles) where the bigger scattered cluster (pink polygon) mainly contains random transitions (red dots). Graph axes are the 1$^{st}$ and 2$^{nd}$ dimensions of Principal Component Analysis(PCA). We used these coordinates to show data points along their maximum variability extension to achieve a clearer visualization. Adjusted Rand Index [24] shows 0.468 agreement between clustering results and real classes.

## Experiment II: CICT predicts direction of associations

We hypothesized that if the composition of transitions contains information about causality, it should be able to predict the direction of causation in the bidirectional association between pairs of clinical conditions. For example, if our observations show frequent both way transition between flu and pneumonia we expect a causal inference method to specify which of the two conditions is the cause or precipitating factor for the other. We used logistic regression and RF to predict the direction of causation in a bidirectional association. Using the ground truth established in experiment I, we selected a set of 225 causal transitions (e.g. flu → pneumonia) and 225 reverse of causal relations (e.g. pneumonia → flu) (Supplementary Table 3). We then used a 75% random sample of this data for training and used the 25% remaining to test the model. We used 10-fold cross-validation on the training set and conducted training of each model 50 times to ensure model stability. RF surpassed logistic regression. Best results achieved with RF using 30 trees with depth 5 and show a discrimination power of AUC= 0.772 with Mean Square Error = 0.193 and R2 = 0.215 on an outstanding validation set. The model was well calibrated across 10 deciles of risk (Hosmer-Lemeshow Chi_square = 5.195, P-value = 0.736). (Figure 2 E, F, G). Top 10 cause-effect predicted relationships in bi-directional transitions, shown in Table 1.B, are well known causal relations in medicine. An AUC = 0.772 is an acceptable discrimination power as a model with an AUC of more than 0.7 is considered as a fair model with practical applications[25].

Figure 2 H shows two clusters, as identified by Partitioning Around Medoids(PAM) [23]. The denser cluster (cyan polygon) mostly embodies cause-effect transitions (blue triangles) where the bigger scattered cluster (pink polygon) mainly contains effect-cause relationships (red dots). Graph axes are the 1st and 2nd dimensions of PCA and axis label shows the variability of data explained by each dimension. Adjusted Rand Index shows[24] 0.437 agreement between clustering results and the real classes reflecting the fact that even unsupervised algorithm can discriminate the direction of causality in association relationship considerably, using CICT features.

## Experiment III: CICT performs well on a random subset of transitions

To ensure that our results are not affected by design decisions in this experiment, we first empirically optimized training a predictive model using a set of 250 cause-effect relations and 90 effect-cause as the positive class, plus 840 random relations as the negative class. Then we used all transitions with frequency > 20 among 873,761 total observed transitions, to create a random sample of size 1600. Next, we used a trained RF model to predict whether each of sampled transitions is on a causal pathway or not. We used predicted value as model's measure of causality. For transitions that occurred in both directions in the results (like A→ B and B→ A), we retained transitions with higher predictive value. Next, we removed any transition with a predicted probability less than a threshold 0.535 and returned 75 relations. The choice of threshold made by applying Youden-Index[26] on prediction results of Experiment II to find the optimal cut-off point. This optimal threshold represents the best performance of discrimination when both effect-cause and cause-effect transitions exist. Next, we asked two clinicians to evaluate the output. CICT did not report any random transition. Among the transitions identified, after removing 13 unexplainable relations due to coding or label ambiguity (e.g. CHF → CHF nonspecific), 62 causal transitions remained. Of these, 52 (p=0.764) were cause-effect and 10 (p=0.147) were effect → cause relations. The top 10 predicted relations are shown in Table 1C. We conducted this experiment 5 times on different random sets of transitions with little variation on prediction accuracy.

## Experiment IV: CICT works independently of cohort and coding structure

One important objective of our study was to create a context independent method that works across datasets and scientific disciplines. Therefore we designed an experiment to understand whether causality can be found regardless of changes in the dataset (as a matter of study design) and coding structure (e.g. ICD9-CM) (as a matter of human intervention on labeling). Accordingly, we defined a cohort of patients with chronic heart failure (CHF, n= 211,284 patients with 1,758,466 admissions) based a previously published definition[27] using data from the California State Inpatient Dataset from 2008-2011. Then in order to label patients' transition events, we used a coding system that groups ICD9-CM codes into a smaller number of clinically meaningful categories, named Clinical Classification Software (CCS) Coding[28]. CCS contains 259 diagnostic code groups, therefore, the resulted graph contained 260 vertices (one for death) and 19890 transition edges. Then for each transition edge, we calculated level 1,2, and 3 features as explained in the method section. Next, two clinicians as subject matter experts labeled a total of 322 transitions including 'causal' and 'not causal' relations for training and testing. Then to predict a binary outcome of causal or not causal, we used a 60% random sample of the 322 relations for training an RF model and used the remaining 40% to test the model. We used 10-fold cross-validation on the training set and we repeated the experiment 50 times to ensure model stability. RF predictive model on average shows a discrimination of AUC=0.831 with Mean Square Error = 0.094 and $R2$ = 0.73. In this experiment, we changed the coding system and created a subset of data by limiting to those with heart failure. The results show that CICT can capture causality regardless of the dataset and event identification process, as long as it is consistent and reflective of the reality. Also, it shows that a uniform change in probabilities by preconditioning on specific states(CHF), did not reduce the amount of causality information that the system learned.  Also in the secondary experiment, we used RF to classify the type of transitions into four groups: (1) Causal, (2) Early and late effect of a common cause, (3) Coexistence and (4) Random. As an example of an interesting clinical finding, CICT finds it significant that atherosclerosis precedes calculus of the kidney which in turn precedes an acute myocardial infarction(AMI). Clinically it means that the calculus of the kidney can be an early warning sign for an elevated risk of an acute myocardial infarction. A literature review to investigate this finding retrieved a meta-analysis of 6 recent cohort studies that confirms an association between calculus of kidney and increased the risk of adverse cardiovascular events including AMI[29] without mentioning the order of events.  Also, early results of this experiment showed that heart failure can be a causative factor for breast cancer and non-Hodgkin lymphoma. We discussed this result with expert clinicians and the group concluded that it is cancer that may induce heart failure because of chemotherapy and radiotherapy (for example anthracycline induced cardiomyopathy). However, a prospective study published[30] three months afterward reported for the first time that heart failure increases the risk of developing cancer. Supplementary table 4 represents 30 interesting inferred causal associations that CICT reports.

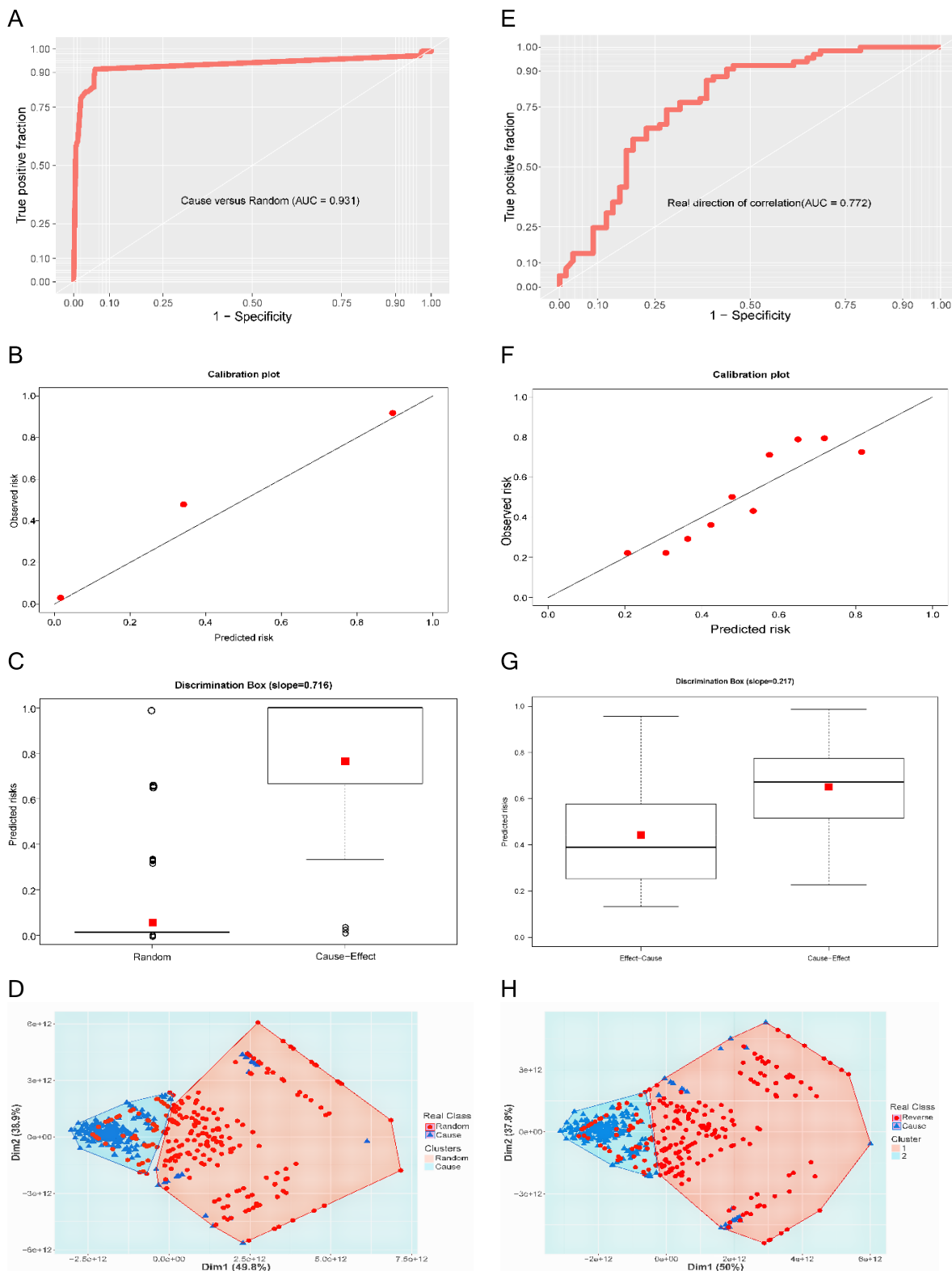

**Fig 2 Left Column:** CICT shows high accuracy in discriminating random transitions from associations. (A) ROC curve. (B) Calibration plot. (C) Discrimination box plot (D) Two clusters as identified by Partitioning Around Medoids along with the real class of data points. **Right Column:** CICT performs well in identifying direction of association: (E) ROC curve. (F) Calibration plot. (G) Discrimination box plot (H) Partitioning Around Medoids
12

### A  Top 10 predicted relationships in experiment I.

| Measure | Source clinical condition | Target clinical condition |
|---|---|---|
| 0.996 | Hyperparathyroidism | Disorders of calcium metabolism |
| 0.996 | Peritonitis NOS | Abdominal pain |
| 0.996 | Pressure ulcer | Bacteremia |
| 0.996 | End stage renal disease | Anemia Not otherwised specified(NOS) |
| 0.996 | Polycystic ovaries | Overweight and obesity |
| 0.996 | Human immunodeficiency virus [HIV] disease | Bacteremia |
| 0.996 | Shock without mention of trauma | Transient alteration of awareness |
| 0.996 | Alcoholic cirrhosis of liver | Portal hypertension |
| 0.996 | Infectious mononucleosis | Splenomegaly |
| 0.996 | Calculus of ureter | Renal colic |

### B  Top 10 Predicted relationships in experiment II

| Measure | Source clinical condition | Target clinical condition |
|---|---|---|
| 0.908 | Systemic lupus erythematosus | Renal failure NOS |
| 0.906 | Hepatic encephalopathy | Transient alteration of awareness |
| 0.887 | Other pyelonephritis or pyonephrosis, not specified as acute or chronic | Renal failure NOS |
| 0.858 | Calculus of ureter | Leukocytosis NOS |
| 0.845 | Systemic lupus erythematosus | Thrombocytopenia NOS |
| 0.815 | Irritable bowel syndrome | Chronic pain |
| 0.803 | Meckel's diverticulum | Unspecified intestinal obstruction |
| 0.798 | Other specified disorders of circulatory system | Edema |
| 0.777 | Vesicoureteral reflux unspecified or without reflux nephropathy | Urinary tract infection, site not specified |
| 0.769 | Septicemia due to other gram-negative organisms | Shock without mention of trauma |

### C  Top 10 predicted relationships in experiment III

| Measure | Source clinical condition | Target clinical condition |
|---|---|---|
| 1.00 | Unspecified hypertensive heart disease without heart failure | CHF NOS |
| 1.00 | Other and unspecified rheumatic heart diseases | CHF NOS |
| 1.00 | Other primary cardiomyopathies | CHF NOS |
| 1.00 | Overweight and obesity | Localized adiposity |
| 1.00 | Mitral valve disorder | CHF NOS |
| 1.00 | Secondary malignant neoplasm of retroperitoneum and peritoneum | Malignant neoplasm of ovary |
| 1.00 | Infection and inflammatory reaction due to internal prosthetic device, implant, and graft | Acquired deformities of hip |
| 1.00 | Malignant essential hypertension | Hypertension NOS |
| 1.00 | Diabetes mellitus complicating pregnancy, childbirth, or the puerperium | Abnormal glucose tolerance of mother, complicating pregnancy, childbirth, or the puerperium |
| 1.00 | Unspecified intracranial hemorrhage | Intracerebral hemorrhage |

**Table 1:** (A) Top 10 predicted association relationships in a mixture of random and causal transitions. For 9 out of 10 the causal direction is predicted correctly. (B) Top 10 predicted direction of causality in bi-directional association transitions (C) Top 10 predicted cause-effect relationship in a random unknown set of transitions.  NOS stands for "Not Otherwise Specified"

## Important predictors of causal relations

Here we evaluate model variables to understand which of the over 300 computed features are important predictors able to discriminate a causal relation from a random one. We used the model in Experiment I and calculated relative importance (RI)[31] of variables on random forests to rank predictors. Then we kept top 8 predictors of the model with a relative importance between 1.0 and 0.043 (figure 3 A). Figure 3 B represents histogram and density graphs of the top predictors in log scale and shows the distribution of predictors for causal and random transitions are results of two different generative processes. We evaluated the significance of differences by non-parametric two-sample Kolmogorov-Smirnov test[32] (p-value for top 6 predictors tends to zero, for intvl_median $p < 0.001$ ).

The most important predictor was scfNMAD.x (Fig 3) which measures the median absolute deviation of normalized confidences of outputs from the source. The median of the distribution of scfNMAD for the source of causal edges (dashed red line), is 3 order of magnitude larger than of random edges (dashed blue line). This suggests that after adjusting for target probabilities, the probability of target conditioned on the source is higher for causal relations. The interesting observation is that without adjusting confidence = P(target | source) for the frequency of target, a simple conditional probability is insignificant (RI<0.001) and cannot differentiate signal from noise.

The second predictor is scbMedian.y which measures the median of contributions of the target to other nodes. We defined contribution as the probability of being previously in a specific primary state once we are in a secondary state. For example, knowing that a patient has pneumonia, what is the probability that he had influenza beforehand? Judging the distribution of scbMedian.y by their medians (vertical dashed red and blue lines), it is one order of magnitude lower for targets of causal transitions comparing to random transitions. This means that effects (targets of a causal transition) usually have a lower rate of contribution to a wider range of others comparing to targets of random transitions. A plausible interpretation is that effects do not contribute significantly to others, so transitions from them to other events tends to be random. This observation characterizes the effect behavior in in a causal relation.

The third predictor ocbNMAD.y, which measures the median absolute deviation of the normalized contribution of nodes into the target, is significantly higher for causal transitions. It means that after adjusting for the source prevalence, on average the influence of inputs into the target of causal transitions is higher than in random transitions.

ScfL1.y which is the 4th predictor means the L-mean of confidences of transitions from target to other nodes. The median of the distribution of ScfL1.y is lower for random relations than for causal relations. This suggests that a random target on average transits to more conditions at lower rates comparing to an effect target which transits to a lower number of conditions each with higher confidence. The distribution demonestrates that effect nodes show a wide range of transition behaviors. One interpretation is that some of the effects act as sinks and modulators that put patients on common afterward care pathways.

Interestingly, we can see that causal transitions have a wider range of time intervals comparing to random transitions judging based on intvl_median: distribution of medians of time intervals of each transition edge. The distribution shows that causal transition may happen as soon as one day or as late as 1700 days with a median of 67.5 days. But random transitions are mostly happening after 10 days with a median of approximately 150 days. The graph of intvl_median shows that the median of intervals for causal transitions is higher than the median of intervals of random transitions.

Another interesting finding is that 6 out of 8 top important predictors are related to node characteristics and just 2 low-rank predictors 'tz',(RI = 0.18) and intvl_median(RI= 0.08) are features of the specific edge. A reasonable interpretation is that some phenomena are by their very nature causal events and some are effects regardless of any other circumstances. Also, it is noteworthy that none of the features that we created to measure the asymmetry of bidirectional transitions, showed in important predictors of causality. Accordingly, the most important factor in determining whether a specific transition is causal or random is the nature of source and target of the transitions.

Evidently, a considerable amount of knowledge about nature of each phenomenon can be gained from the composition of its previous (input) and afterward (output) events. Also, it is the nature of source and target that largely specifies the type of transition between them. We can conclude that to understand whether a specific transition is causal or random depends on a higher order or meta structure of inputs and outputs to source and target. This results in three important findings: (1) standard Markov chains contain implicit hidden structure that is richer in information than it is previously known; (2) considering that some of the important predictors are confidences of other nodes into source of an edge and contribution of target into other nodes suggests that beyond one step transitions a higher level or meta-structure exist in memoryless Markov chains which deserve further exploring; and (3) analysis of composition of input and outputs can reveal important causal pathway in prevalent, real life, one step transition networks.

### CICT Time and space complexity

For training and prediction steps CICT has a space-time complexity like the specific predictive model used. For example for random forests, it is $O(e^* \log_e)$ for training on a subset of edges 'e' which is usually a small subset of the whole edge sets and $O(E)$ for prediction over the set of all edges E.

(A) Important discriminating predictors of Causal versus Random relations

|   | Variable | scaled importance | Description |
|---|----------|-------------------|-------------|
| 1 | scfNMAD.y | 1.000 | Median absolute deviation of normalized confidences from target |
| 2 | scbMedian.x | 0.977 | Median of source's contributions into other nodes |
| 3 | ocbNMAD.x | 0.402 | Median absolute deviation of normalized confidence of other into source |
| 4 | scfL1.x | 0.367 | Mean of confidences of source using L-Measures |
| 5 | PinSD.y | 0.272 | Standard deviation of power of inputs into the target |
| 6 | Tz | 0.184 | Sum of Z-score of confidence and contribution of edge |
| 8 | ocfKurt.x | 0.090 | Kurtosis of others contribution into source |
| 9 | intvl_median | 0.086 | Median of intervals all observed transitions between source and target |

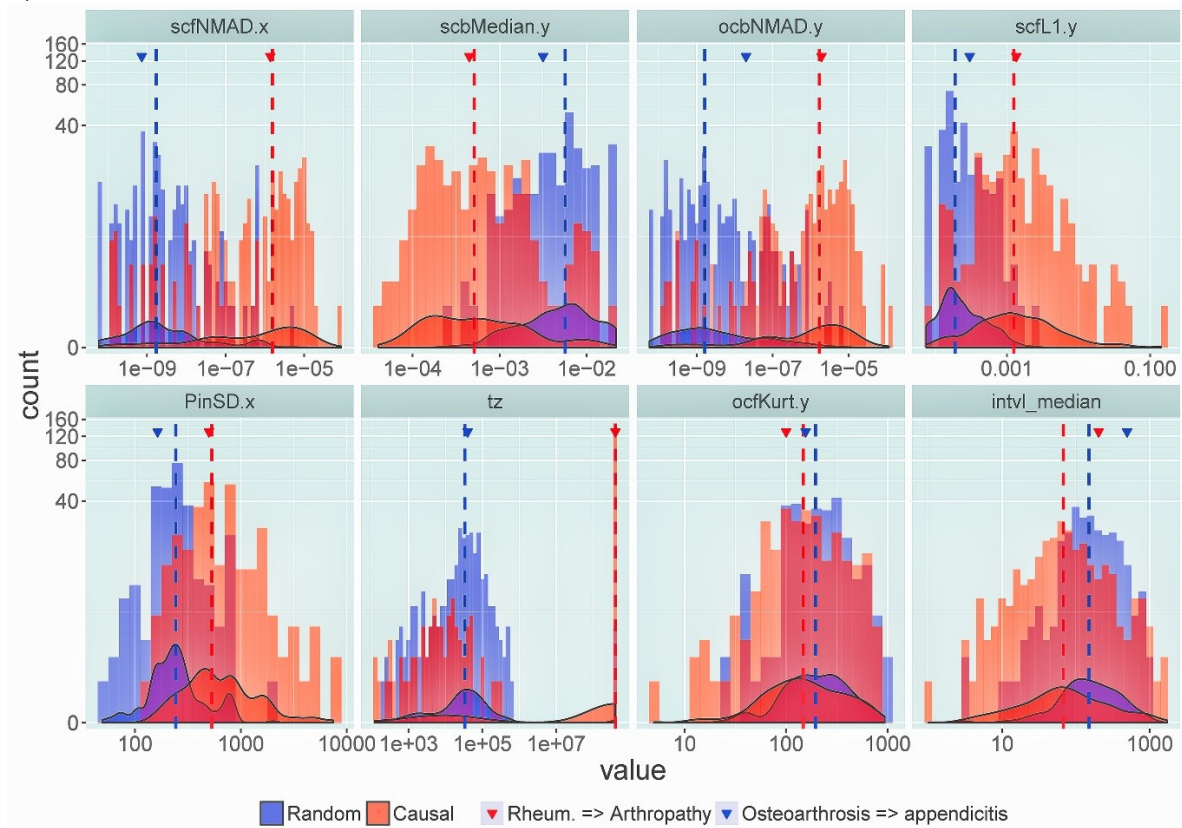

Figure 3: (A) Important discriminating predictors of Causal versus Random relations (B) Distribution of most important discriminating predictors of causal transitions (red) from random ones (blue). A logarithmic scale used for x and y. Histogram and density area graphs are superimposed to better reflect differences in distribution. Vertical dash lines show the median of the distribution. Red point shows the location of a causal relation 'Rheumatoid arthritis → Arthropathy, unspecified, pelvic region and thigh" and blue triangle shows the location of irrelevant transition 'Osteoarthrosis → Acute appendicitis". The significant differences of distribution suggest that they are the result of different generative processes, confirmed by two-sample Kolmogorov-Smirnov test.

## Discussion

In the causality inference literature, it is traditionally thought that short-term data cannot provide enough information to infer the causal relation[33] and almost all data-driven causality inference methods need time-series data of sufficient length (usually more than 25 points).

Here, we introduced causality inference using composition of transitions (CICT) as a novel and general analytic method for causality inference and complex system identification in Markov chain data(MC). MC data is frequent in many real-world scenarios in different disciplines where only short-term one-step transition data exist. These complex scenarios happen frequently, such as in econometrics or high throughput biological data[34], also in physics, web page ranking, molecular and higher order phenotypes, and epidemiology[15]. Network identification in MC data is generally considered as a hard and computationally expensive problem due to the exponential increase of candidate networks given the number of nodes. In such scenarios, CICT can reveal the underlying system or dependency structure efficiently. Importantly CICT is free from constraints on the network structure (e.g. acyclic structure) or data structure (e.g. separable cause and effect) or by modeling assumptions (e.g. existence of sufficient length of time series, or following specific distributions). Being free from these three groups of constraints CICT can be applied in a range of contexts and for various objectives and in combination with existing causal inference methods. Also, we showed that the asymmetry of back and forth transitions between two events are less important than distributional features. This makes CICT equally applicable to undirected graph network structures or where a concept of time does not exist such as correlation networks. Due to its simplicity, efficiency and generalizability, CICT has a potential influence on applications and analysis of Markov Chain data across disciplines.

The idea of using compositions of inputs/outputs in CICT introduces a new and rich set of features and reveals some previously unknown facets of causality. We showed that distributional facets of causality continue to persist even when the observations are filtered by preconditioning or when we change the grouping of events, as far as the process is consistent and reflective of real phenomena. For example, in experiment **III** we chose a different cohort, used a precondition to filter and chose a subset of data and we used CCS coding which groups all clinical conditions into 255 events instead of over 10000 ICD9-CM groups that were used in three other experiments. However, these changes had little effect on prediction power of the method. Also, we suggest that CICT is resilient against adding or dropping parts of information as it extracts features from stable measures of distributions like median absolute deviation[17,18] and L-moments[20,35]. This makes CICT robust to unmeasured or latent confounding factors. Another advantage of CICT is that it allows the utilization of random sampling methods and statistics methods for identifying distribution parameters to reduce computation. This quality along with low time-space complexity, allows CICT to be used in the analysis of massive and dense graph data.

From a healthcare research point of view to the best of our knowledge, this study is the first to describe methods to drive broad causal inference using administrative data which has been considered unfit for causal inference due to low clinical content and coding errors. Another significance of our method is the departure from the conventional experimental or observational study design paradigm for identifying and measuring correlations and causal relations in healthcare. In their seminal paper "Causation and causal inference in epidemiology" K. J. Rothman et al state, "Philosophers agree that causal propositions cannot be proved, and find flaws or practical limitations in all philosophies of causal inference. Hence, the role of logic, belief, and observation in evaluating causal propositions is not settled. Causal inference in epidemiology is better viewed as an exercise in the measurement of an effect rather than as a criterion-guided process for deciding whether an effect is present or not"[36]. Despite their limitations, observational studies are often the only way to address many important causal questions[37]. Thus, observational studies are a necessary part of our causal toolbox. Here we show how observational data provides the simple transition rates between clinical conditions and carry valuable information to reveal causal relations even without using contextual information such as age, gender, race or clinical factors.
.

The possibility of identifying causal networks from their compositional behavior reveals new facets of causality and provides another tool for system identification in the frequently available and low-cost Markov Chain data. Moreover, it has implications for our understanding of causality. As a future topic, we will seek to apply CICT in other domains and will consider combining our method with existing causality inference methods to enhance their performance.

35  Karvanen, J. Estimation of quantile mixtures via L-moments and trimmed L-moments. *Computational Statistics & Data Analysis* **51**, 947-959, doi:http://dx.doi.org/10.1016/j.csda.2005.09.014 (2006).
36  Rothman, K. J. & Greenland, S. Causation and Causal Inference in Epidemiology. http://dx.doi.org/10.2105/AJPH.2004.059204, doi:S144.pdf (2011).
37  Reiter, J. Using statistics to determine causal relationships. *The American Mathematical Monthly* **107**, 24-32 (2000).

**Code Availability:**  The required code for training and testing machine learning methods used in this research and for reproducing the results will be available on Github and on a specific website that is under design for this project.

**Data Availability:** The patient transition network datasets generated during and/or analyzed during the current study and the Ground Truth tables are available on '*figshare*' repository.

**Acknowledgments:** We thank Harlan Krumholz, Ronald Coifman, Shu-Xia Li, Paul Horak, Andreas Coppi and Kumar Dharmarajan for their comments and Helen Arjmandi for visualizations and assistance with this work. This research is supported by the Center for Outcomes Research and Evaluation, Yale University.

# Figure Legend

### Figure 1: General concepts and data used for CICT method

(**A**) Descriptive statistics on the data. (**B**) A set of transitions for four hypothetical patients. For example, the first patient is hospitalized with a principal diagnosis of condition X, and after a period is rehospitalized with condition Y and so forth. If we start merging similar transitions the result would be the transition graph shown in (**C**) Different types of transitions on a network. X is a common cause of Y and Z where Y is an early effect and Z is a late effect. X and W showed an association without an observable causal relation. Numbers represent hypothetical frequencies. (**D**) 4 zones that carry different distributional information. It is important to note that the 4 areas are not overlapping and contain different information. Here **i** represents source and **j** represents destination. (**E**) The log-scale density graph shows the different distributions in four distribution zones for a cause: Rheumatoid Arthritis(red), an effect: Syncope (yellow) and a random event: Pneumonia (blue). (**F**) The 8 distribution zones that are identified above carry information relevant to the nature of the transition between source and destination. Zones 1,3,5,7 capture distribution of the parameters that are derived from Confidence calculation. Zones 2,4,6,8 capture distribution of the parameters that are derived from Contribution calculation.

### Figure 2: Evaluation results of Experiments using CICT

**Left Column:** CICT shows high accuracy in discriminating random transitions from associations. (**A**) ROC curve. (**B**) Calibration plot. (**C**) Discrimination box plot (**D**) Two clusters as identified by Partitioning Around Medoids along with the real class of data points. **Right Column:** CICT performs well in identifying direction of association: (**E**) ROC curve. (**F**) Calibration plot. (**G**) Discrimination box plot (**H**) Partitioning Around Medoids

### Figure 3: Top 8 important predictors of causal transitions

(**A**) Important discriminating predictors of Causal versus Random relations (**B**) Distribution of most important discriminating predictors of causal transitions (red) from random ones (blue). A logarithmic scale used for x and y. Histogram and density area graphs are superimposed to better reflect differences in distribution. Vertical dash lines show the median of the distribution. Red point shows the location of a causal relation 'Rheumatoid arthritis → Arthropathy, unspecified, pelvic region and thigh" and blue triangle shows the location of irrelevant transition 'Osteoarthrosis → Acute appendicitis". The significant differences of distribution suggest that they are the result of different generative processes, confirmed by two-sample Kolmogorov-Smirnov test.